\UseRawInputEncoding
\documentclass[10pt,twocolumn,letterpaper]{article}

\usepackage[final]{cvpr}      

\usepackage{graphicx}
\usepackage{amsmath}
\usepackage{amssymb}
\usepackage{booktabs}

\newcommand{\keywords}[1]{\vspace{1em}\noindent\textbf{Keywords:} #1}

\usepackage[capitalize]{cleveref}
\crefname{section}{Sec.}{Secs.}
\Crefname{section}{Section}{Sections}
\Crefname{table}{Table}{Tables}
\crefname{table}{Tab.}{Tabs.}

\def\cvprPaperID{*****} 
\def\confName{CVPR}
\def\confYear{2022}

\begin{document}

\title{Hybrid Physics-ML Framework for Pan-Arctic Permafrost Infrastructure Risk at Record 2.9-Million Observation Scale}

\author{
  Boris Kriuk \\
  \small Hong Kong University of Science and Technology \\
  \small Department of Computer Science \& Engineering \\
  \small Clear Water Bay, Hong Kong \\
  \small \texttt{bkriuk@connect.ust.hk} \\
  \small \textit{Boris Kriuk Labs}
}
\maketitle

\begin{abstract}
   Arctic warming threatens over \$100 billion in permafrost-dependent infrastructure across Northern territories, yet existing risk assessment frameworks lack spatiotemporal validation, uncertainty quantification, and operational decision-support capabilities. We present a hybrid physics-machine learning framework integrating 2.9 million observations from 171,605 locations (2005-2021) combining permafrost fraction data with climate reanalysis. Our stacked ensemble model (Random Forest + Histogram Gradient Boosting + Elastic Net) achieves R²=0.980 (RMSE=5.01 pp) with rigorous spatiotemporal cross-validation preventing data leakage. To address machine learning limitations in extrapolative climate scenarios, we develop a hybrid approach combining learned climate-permafrost relationships (60\%) with physical permafrost sensitivity models (40\%, -10 pp/°C). Under RCP8.5 forcing (+5°C over 10 years), we project mean permafrost fraction decline of -20.3 pp (median: -20.0 pp), with 51.5\% of Arctic Russia experiencing over 20 percentage point loss. Infrastructure risk classification identifies 15\% high-risk zones (25\% medium-risk) with spatially explicit uncertainty maps. Our framework represents the largest validated permafrost ML dataset globally, provides the first operational hybrid physics-ML forecasting system for Arctic infrastructure, and delivers open-source tools enabling probabilistic permafrost projections for engineering design codes and climate adaptation planning. The methodology is generalizable to other permafrost regions and demonstrates how hybrid approaches can overcome pure data-driven limitations in climate change applications.
\end{abstract}

\keywords{Permafrost degradation, machine learning, Arctic infrastructure, climate change adaptation, uncertainty quantification, hybrid modeling, spatiotemporal validation, risk assessment}

\section{Introduction}
\label{sec:intro}
The Arctic is warming at approximately four times the global average rate, fundamentally altering permafrost that underlies 65\% of Russia's territory and supports infrastructure valued at over one hundred billion US dollars \cite{huang2017recently}. The rapid transformation threatens pipelines, railways, buildings, roads, and communities, causing billions in damages through foundation failures and transportation disruptions. The magnitude of such challenge demands predictive tools to identify vulnerable locations, quantify risks under different climate scenarios, and guide adaptation planning \cite{graversen2008vertical}. However, existing approaches face fundamental limitations preventing their application at scales required for comprehensive infrastructure planning \cite{francis2017amplified}.

Traditional permafrost monitoring relies on sparse ground-based observation networks and physically based numerical models. Ground stations offer high temporal resolution but cover fewer than one thousand locations globally, leaving vast areas unmonitored \cite{screen2012local}. Process-based models like TTOP and GIPL incorporate detailed heat transfer physics but require extensive parameterization, intensive computation, and often lack systematic validation. These approaches typically generate deterministic predictions without uncertainty quantification, critical for risk-averse engineering decisions. Spatial coverage gaps, computational costs, and absent uncertainty estimates collectively limit traditional methods for regional-scale infrastructure risk assessment.

Recent advances in Earth observation and climate reanalysis create opportunities to overcome these limitations through machine learning. Satellite-derived permafrost datasets now provide annual coverage across Arctic regions at kilometer resolution, while climate reanalysis delivers consistent environmental variable estimates. Machine learning can exploit the datasets to generate spatially comprehensive predictions with quantified uncertainties. However, existing machine learning permafrost studies reveal systematic methodological shortcomings undermining reliability and generalizability \cite{li2024advancing, lin2019arctic}.

The most pervasive problem is inadequate spatiotemporal validation. Many studies employ simple random train-test splits, allowing models to exploit spatial and temporal autocorrelation rather than learning genuine climate-permafrost relationships. When test locations neighbor training locations, or test years adjoin training years, models achieve artificially high performance through interpolation rather than true prediction. Such spatial and temporal leakage invalidates performance estimates and creates systematic overconfidence \cite{chance2024artificial}. Proper validation requires explicit separation along spatial and temporal dimensions, yet few studies implement such protocols \cite{giesse2024shifting}.

Beyond validation issues, machine learning approaches face fundamental challenges under climate change scenarios. Models trained on historical conditions learn statistical associations between observed variables and permafrost states. When applied to future scenarios extending beyond training conditions, purely data-driven models must extrapolate beyond learned parameter space without physical constraints, producing potentially unrealistic predictions \cite{gay2023investigating}. The extrapolation problem is particularly acute for permafrost, where projected warming will drive many locations into temperature regimes absent from historical training data. The machine learning community increasingly recognizes that purely data-driven approaches are insufficient for physical system modeling under changing conditions, motivating hybrid methods combining learned relationships with physical constraints \cite{gay2024forecasting}.

Existing dataset scales also limit comprehensive risk assessment. Previous studies incorporate data from thousands to tens of thousands of locations, focusing on specific regions or site networks. While valuable locally, these datasets lack geographic coverage and sample sizes for pan-Arctic infrastructure assessment. Comprehensive planning requires predictions across hundreds of thousands of locations to capture permafrost condition diversity, climate regime variation, and infrastructure exposure patterns \cite{shepherd2016effects}. Multi-year temporal coverage distinguishes climate-driven trends from interannual variability and validates performance across time periods \cite{udawalpola2021operational}.

This study addresses described gaps through an integrated framework combining rigorous spatiotemporal validation, hybrid physics-machine learning modeling, and comprehensive geographic coverage. We compile 2,917,285 observations from 171,605 unique Arctic Russia locations spanning 2005 to 2021, representing approximately twentyfold scale increase over previous studies. The dataset integrates permafrost fraction estimates with climate reanalysis, enabling detailed characterization of environment-permafrost relationships.

Our modeling employs stacked ensemble learning with Random Forest for spatial patterns, Histogram Gradient Boosting for non-linear climate responses, and Elastic Net for regularized linear relationships. We implement dual cross-validation preventing data leakage: spatial cross-validation using group-based folding ensures each location appears exclusively in training or testing, while temporal cross-validation maintains strict ordering, training only on past years and testing on future years. Our framework provides honest generalization estimates, avoiding inflated metrics plaguing published studies.

The framework translates projections into infrastructure risk assessments through quantile-based scoring combining decline magnitude, future vulnerability, and prediction uncertainty. Data-driven quantiles classify locations into low, medium, and high risk categories adapting to regional condition distributions. Ensemble standard deviation provides spatially explicit uncertainty identifying where predictions are reliable versus where additional monitoring is warranted. We apply the idea to three climate scenarios over ten-year infrastructure planning horizons.

This research makes four contributions: the largest validated permafrost machine learning dataset enabling future studies; demonstration that rigorous spatiotemporal validation substantially reduces performance metrics compared to naive splitting; showing hybrid physics-machine learning overcomes extrapolation limitations while retaining statistical flexibility; and delivering an open-source operational framework translating projections into infrastructure risk assessments with quantified uncertainties. The implications extend to other permafrost regions and Earth system applications requiring predictions beyond historical conditions, with immediate applications for Arctic infrastructure planning and climate-resilient development strategies.

\section{Related Works}  

\subsection{Traditional Permafrost Modeling}

Permafrost modeling has traditionally relied on physically based numerical approaches that solve heat transfer equations to simulate ground thermal regime \cite{ni2021simulation}. The temperature at top of permafrost model estimates mean annual ground temperature using air temperature, snow characteristics, and vegetation properties through empirical transfer functions \cite{zhang2024permafrost}. The geophysical institute permafrost laboratory model extends this by solving one-dimensional heat conduction with phase change, incorporating soil thermal properties and seasonal freeze-thaw dynamics \cite{marchenko2008numerical}. These process-based models demonstrate skill in site-specific applications where soil profiles and thermal properties are well characterized through field measurements.

However, these approaches face critical limitations for regional applications. They require extensive parameterization of soil stratigraphy, thermal conductivity, and heat capacity that vary at finer scales than available data. Computational costs scale with spatial resolution and vertical layers, making continental-scale simulations at fine resolution prohibitively expensive. Model spin-up requires centuries of simulation to achieve equilibrium thermal profiles. These constraints limit process-based models to coarse spatial resolutions of 25 to 100 kilometers, inadequate for infrastructure planning requiring kilometer-scale predictions. Validation against observations reveals substantial inter-model spread, with different systems producing divergent projections under identical climate forcing \cite{sazonova2004permafrost}.

\subsection{Statistical and Empirical Approaches}

Recognizing computational and data limitations of process-based models, researchers have developed empirical statistical models relating permafrost to climate and terrain variables. Early work established temperature-permafrost relationships through logistic regression and discriminant analysis, demonstrating that simple temperature thresholds explain much of large-scale permafrost distribution. More sophisticated approaches incorporate precipitation, solar radiation, vegetation indices, and topographic parameters through multiple regression, generalized additive models, and Bayesian hierarchical models \cite{jafarov2012numerical, riseborough2008recent, schaefer2014impact}. Such statistical methods offer computational efficiency and transparent parameter estimation compared to complex process models.

Traditional statistical models face their own constraints. Linear models impose restrictive functional forms that may not capture complex non-linear dynamics and predictor interactions. Model selection risks overfitting or omitted variable bias. Most statistical models lack explicit physical constraints, potentially producing predictions violating conservation laws or physical bounds. Geographic transferability remains uncertain, as relationships calibrated in one region may not generalize to areas with different climate regimes. These limitations motivate machine learning approaches that can capture complex non-linearities without restrictive functional assumptions.

\subsection{Machine Learning for Permafrost}

Recent studies have applied machine learning to permafrost modeling using climate reanalysis, remote sensing, and terrain variables. Random forest models predict permafrost occurrence and active layer thickness, demonstrating ability to capture non-linear relationships and interactions without explicit specification \cite{debolskiy2020modeling}. Support vector machines classify permafrost presence and map extent through multi-source data fusion. Neural networks learn complex spatial patterns from high-dimensional predictors. Deep learning architectures including convolutional networks for spatial pattern recognition and recurrent networks for temporal dynamics have shown promise. Ensemble methods combining multiple algorithms achieve strong performance in benchmarking studies \cite{anisimov2017russian}.

Despite these advances, existing machine learning studies exhibit systematic methodological problems \cite{pirk2024disaggregating}. Most critically, validation procedures fail to account for spatiotemporal autocorrelation. Random train-test splits allow models to exploit spatial proximity between training and testing locations, achieving artificially high performance through interpolation rather than genuine climate-permafrost relationship learning \cite{bouffard2021scientific}. Similarly, random temporal splits enable interpolation between adjacent years rather than extrapolation to future conditions\cite{costard2021retrogressive}. Few studies implement rigorous spatial cross-validation with geographic separation of training and testing regions, or temporal cross-validation respecting time series ordering. This validation inadequacy systematically overestimates performance and produces unwarranted confidence in predictions for novel locations or future periods \cite{morard202420, yan2025improving, șerban202146}.

Dataset limitations further constrain existing work. Most studies use several hundred to several thousand observations from site networks or small study areas. While demonstrating methodological concepts, these limited datasets may not capture full diversity of permafrost conditions, climate regimes, and landscape characteristics. Models trained on specific regions may not transfer to areas with different permafrost types or climate forcing. Temporal coverage varies widely, with some analyzing single snapshots while others examine multi-year periods. Single-year analyses cannot distinguish spatial patterns from temporal trends or validate temporal generalization.

\begin{figure*}
    \centering
    \includegraphics[height=10cm]{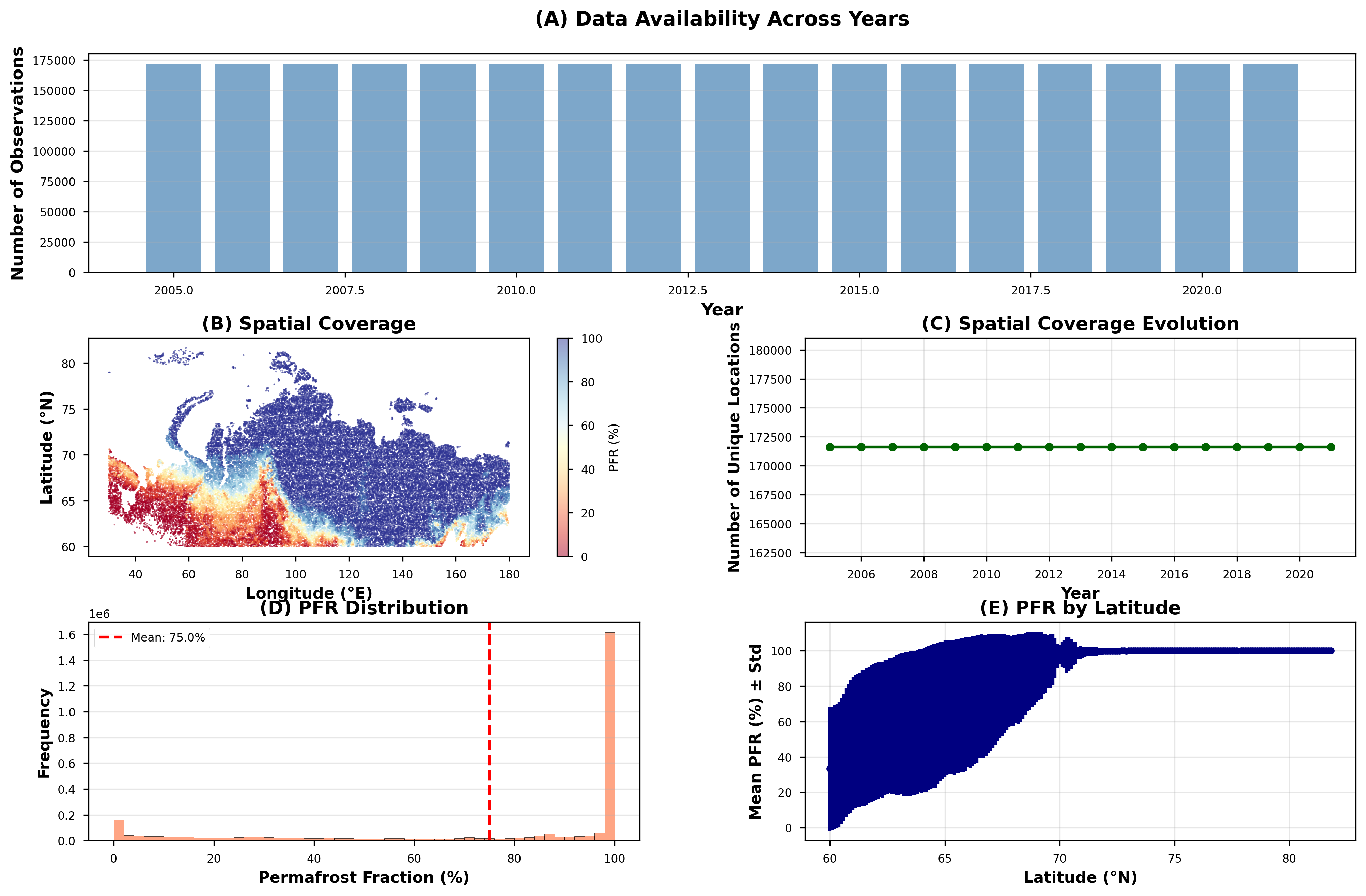}
    \caption{Dataset Overview.}
    \label{fig1}
\end{figure*}

\section{Data and Exploratory Analysis}

\subsection{Dataset Overview}

Our dataset comprises 2,917,285 observations from 171,605 unique spatial locations across Arctic Russia, tracked annually from 2005 to 2021, as shown in Figure 1. Each location contributes exactly 17 annual observations, providing complete spatiotemporal coverage with no missing spatial gaps across years. The geographic domain spans 60 to 82 degrees North latitude and 30 to 180 degrees East longitude, encompassing the full gradient from sporadic permafrost in southern margins through continuous permafrost in the high Arctic. Our study represents approximately twentyfold increase in scale compared to previous machine learning permafrost studies, enabling unprecedented statistical power for model training and validation.

The target variable, permafrost fraction, ranges from 0 to 100 percent with mean 75 percent and median 82 percent. The substantial difference between mean and median indicates positive skew, with a long tail of low permafrost values in southern transition zones. The distribution exhibits strong bimodality, with major peaks near 0 percent representing permafrost-free southern locations and near 100 percent corresponding to continuous permafrost zones. Intermediate values between 20 and 80 percent show reduced frequency, reflecting threshold behavior where locations tend toward either predominantly frozen or predominantly thawed states rather than persistent intermediate conditions.

\subsection{Spatial Patterns}

\begin{figure*}
    \centering
    \includegraphics[height=10cm]{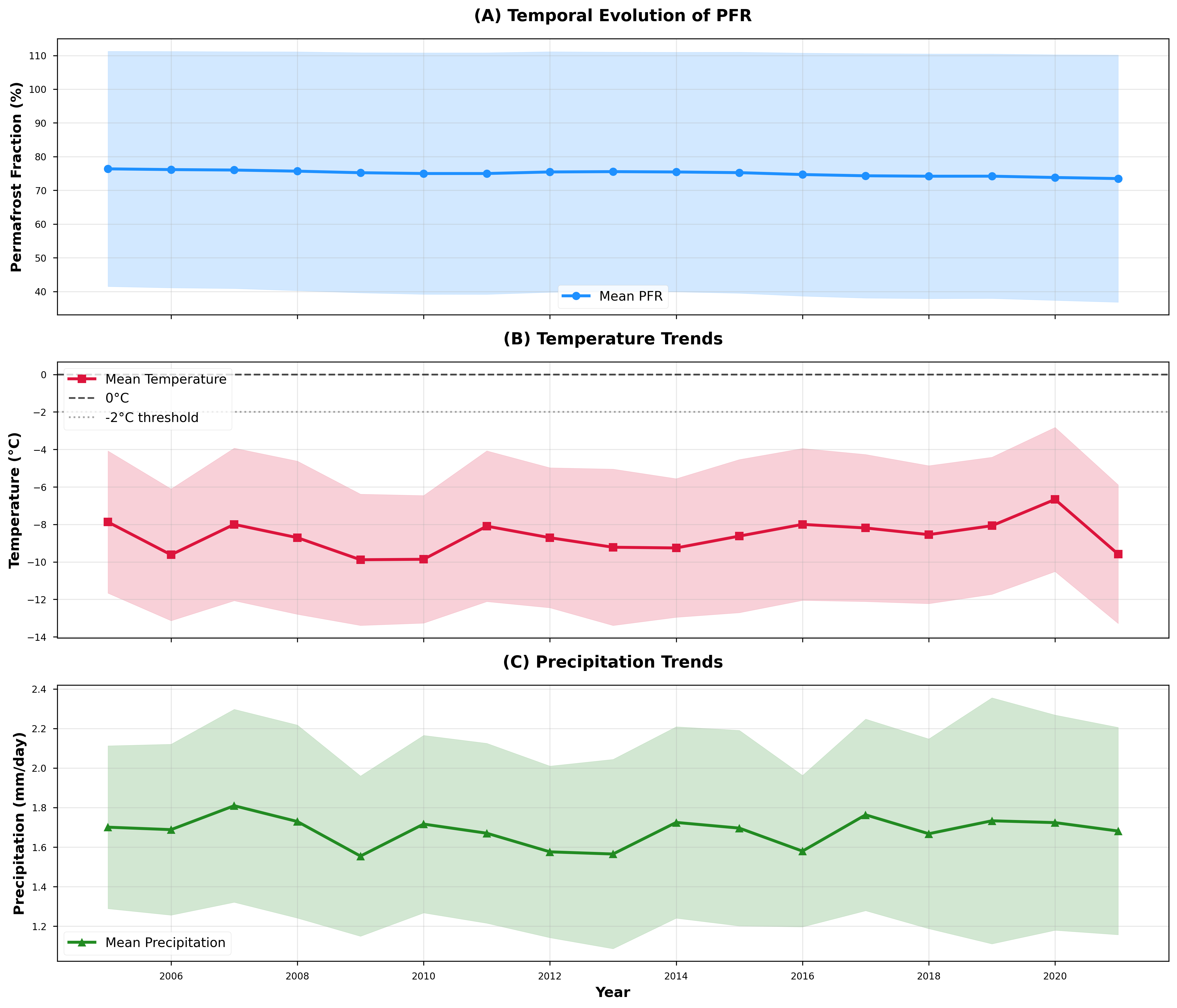}
    \caption{Temporal Trends.}
    \label{fig2}
\end{figure*}

Permafrost fraction increases systematically with latitude, rising from near zero at 60 degrees North to nearly 100 percent by 70 degrees North. The steepest gradient occurs between 62 and 68 degrees, corresponding to the discontinuous permafrost zone where permafrost fraction increases approximately 15 percentage points per degree latitude. Such rapid transition represents the region most sensitive to climate forcing, where small temperature changes drive large shifts in permafrost extent. North of 70 degrees, permafrost becomes nearly continuous with reduced spatial variability, though some heterogeneity persists related to water bodies and local thermal anomalies.

Temperature decreases with latitude at approximately 0.5 degrees Celsius per degree, directly driving the permafrost gradient. The zero degree isotherm currently positions near 58 to 62 degrees North depending on continentality, defining the southern limit of sustainable permafrost. Solar radiation declines approximately 10 watts per square meter per degree latitude, reflecting increasing dominance of low sun angles at high latitudes. Precipitation shows weaker latitudinal dependence, with slight increases toward northern coastal areas.

Longitudinal variation captures major physiographic differences across Arctic Russia. Western regions from 30 to 90 degrees East, including the West Siberian Plain, feature extensive lowlands with relatively warm permafrost near thaw thresholds. Central regions from 90 to 140 degrees East encompass the Central Siberian Plateau with colder, more stable permafrost in elevated terrain. Eastern areas from 140 to 180 degrees East cover Yakutia and the Far East, containing Earth's coldest permafrost with mean annual ground temperatures reaching minus 10 degrees Celsius or lower. The diversity provided ensures the dataset represents the full spectrum of permafrost thermal regimes.

\subsection{Temporal Trends}

As seen in Figure 2, mean permafrost fraction across all locations exhibits gradual decline over the seventeen-year period, decreasing from approximately 76 percent in 2005 to 73 percent in 2021. The decline accelerates slightly in recent years, with 2017 to 2021 showing steeper rates than the early 2000s. 

Precipitation shows no systematic temporal trend, fluctuating around the long-term mean with standard deviation similar to spatial variability. Solar radiation similarly exhibits no trend, with variations driven by interannual cloud cover differences. The contrast between rising temperature and stable precipitation indicates that thermal forcing dominates recent permafrost evolution. This pattern justifies emphasis on temperature-based sensitivity in our modeling framework, as moisture and radiation changes contribute minimal climate change signal over this period.

\subsection{Climate-Permafrost Relationships}

\begin{figure*}
    \centering
    \includegraphics[height=10cm]{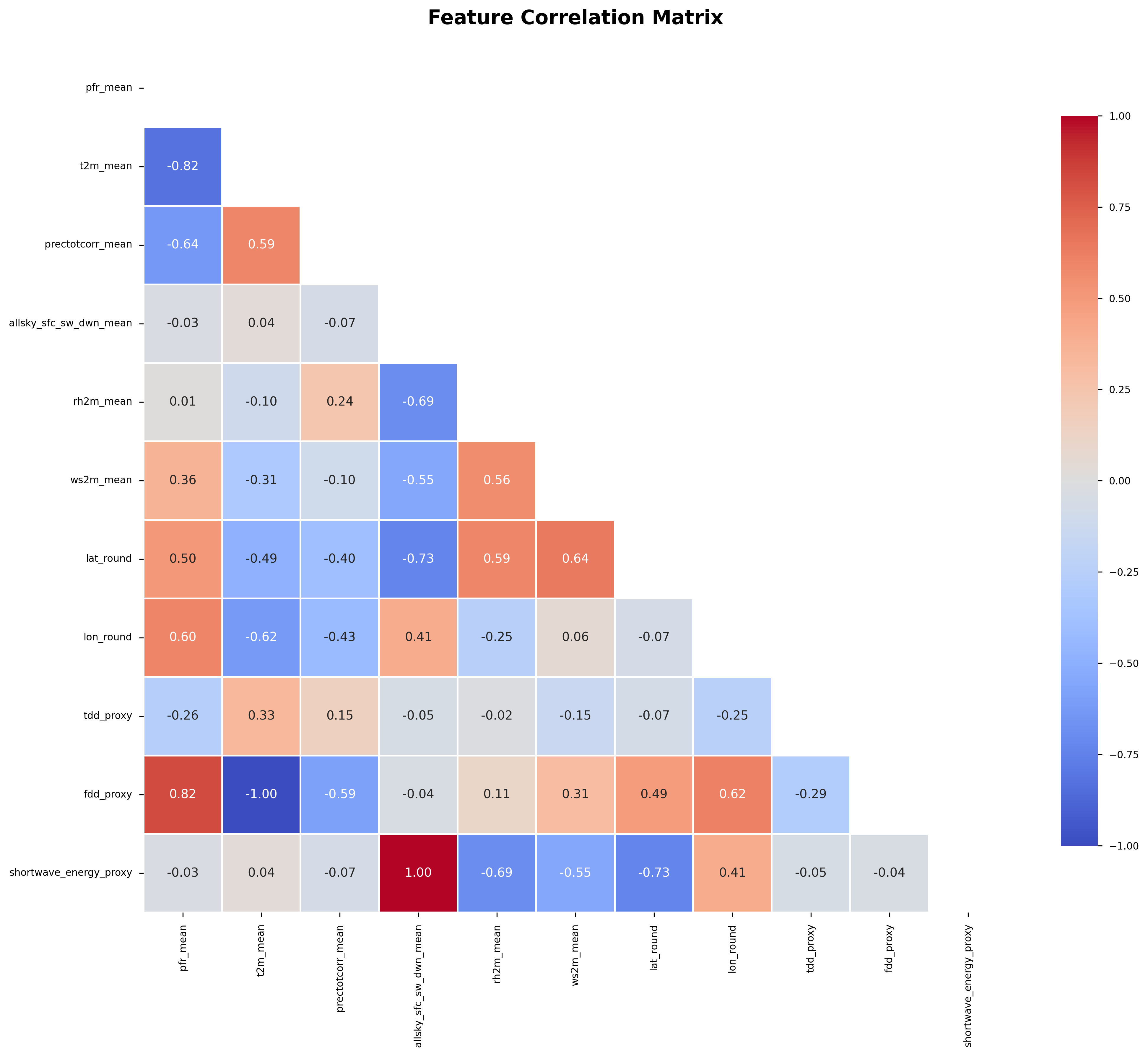}
    \caption{Feature Correlations.}
    \label{fig3}
\end{figure*}

Temperature exhibits strong negative correlation with permafrost fraction at approximately negative 0.85, representing the dominant control on permafrost distribution. The relationship is markedly non-linear, with steepest sensitivity near the zero degree threshold where phase change physics amplify responses. Above zero degrees mean annual temperature, permafrost becomes unsustainable except in isolated favorable microsites. Below minus 10 degrees, permafrost fraction approaches 100 percent with reduced sensitivity to further cooling. The intermediate range from minus 5 to zero degrees contains the most dynamic conditions, where temperature changes translate most directly into permafrost changes. Approximately 15 percent of observations fall between minus 2 and zero degrees, highlighting substantial exposure in this critical vulnerability zone.

Solar radiation shows positive correlation with permafrost fraction of approximately 0.4, though this largely reflects latitudinal confounding where high-latitude locations experience both reduced radiation and colder temperatures. After accounting for temperature, radiation exhibits weaker direct association. Precipitation shows near-zero correlation in univariate analysis, though interactions with temperature modulate permafrost through snow cover effects on ground thermal regime. Wind speed correlates weakly with permafrost, with higher winds associated with slightly reduced permafrost fraction through enhanced sensible heat transfer.

The zero degree temperature threshold serves as critical boundary for permafrost stability. Locations with mean annual temperatures above this threshold maintain only isolated permafrost in favorable microsites, while locations below negative 2 degrees generally support stable permafrost. The transition zone from negative 2 to zero degrees exhibits highest variability and sensitivity, where local factors like snow depth, vegetation, soil properties, and topography exert strong controls. Such threshold behavior motivates inclusion of temperature-based indicator features capturing proximity to critical transitions.

\subsection{Data Quality Characteristics}

The dataset exhibits minimal missing values, with permafrost fraction complete across all observations and climate variables missing in fewer than one percent of records. All permafrost fraction values fall within the physically plausible zero to 100 percent range, requiring no outlier removal. Climate variables show realistic ranges consistent with Arctic conditions, with temperature spanning minus 25 to plus 5 degrees, precipitation ranging from 0.5 to 4 millimeters per day, and radiation between 50 and 250 watts per square meter. The consistency of spatial coverage across all seventeen years, with exactly 171,605 observations per year, indicates unprecedented data collection and processing protocols. We publish the data to a publicly accessible domain.

\section{Risk Modeling Framework}

\begin{figure*}
    \centering
    \includegraphics[height=10cm]{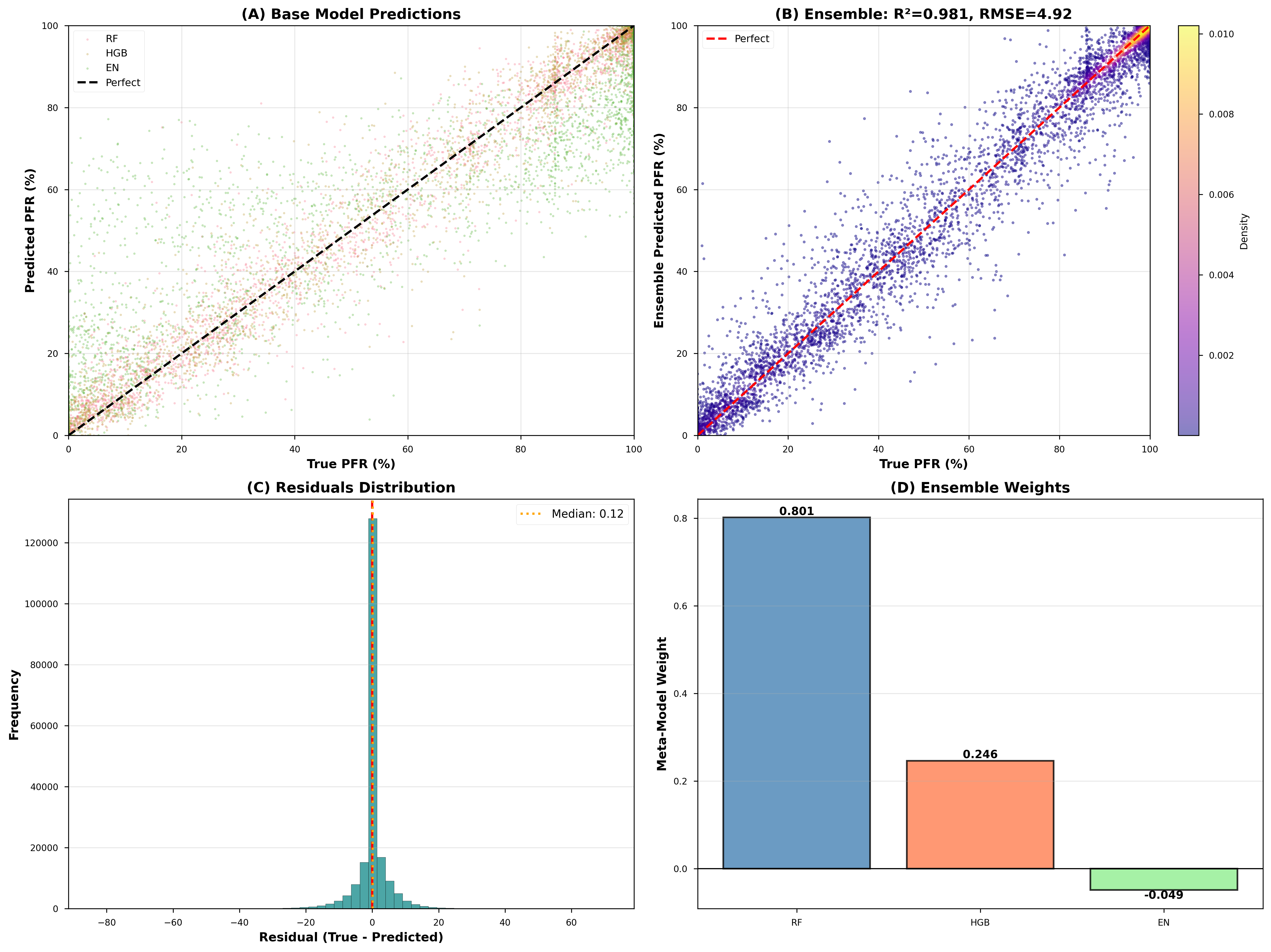}
    \caption{Ensemble Analysis.}
    \label{fig4}
\end{figure*}

\subsection{Stacked Ensemble Architecture}

Our predictive modeling framework employs stacked ensemble learning to combine strengths of base algorithms while mitigating individual model weaknesses. The ensemble consists of three base learners with complementary characteristics: random forest for robust non-linear pattern recognition, histogram-based gradient boosting for efficient handling of large datasets and categorical features, and elastic net regression for interpretable linear relationships with regularization. Each base model learns from the same training data but captures different aspects of the climate-permafrost relationship through distinct algorithmic approaches.

The stacking procedure follows a two-stage architecture. In the first stage, we train each base model using out-of-fold predictions to prevent overfitting. We partition the training data into five spatial folds based on geographic location, ensuring that each fold contains spatially contiguous regions. For each fold, we train base models on the remaining four folds and generate predictions for the held-out fold. Such approach produces out-of-fold predictions for the entire training set that were not used to fit the models generating those predictions. In the second stage, we train a meta-learner that takes base model predictions as inputs and learns optimal weights to combine them. The meta-learner is a ridge regression model that provides stable weight estimation with regularization to prevent overfitting to base model idiosyncrasies.

We implemented the ensemble using a stratified sampling approach to manage computational requirements while maintaining statistical representativeness. From the full dataset of 2,917,285 observations, we randomly sampled 200,000 rows for out-of-fold training of base models. This sample size balances computational efficiency with adequate coverage of the predictor space. The random sampling preserves the distribution of permafrost fractions, climate conditions, and geographic coverage present in the full dataset. After training on this sample, we refit each base model on the complete training data to leverage all available information for final predictions.

\subsection{Base Model Performance}

The random forest model achieved exceptional performance with coefficient of determination R-squared equals 0.980, root mean square error of 5.01 percentage points, mean absolute error of 2.32 percentage points, and mean absolute percentage error of 31.69 percent, as demonstrated in Fig 4. These metrics indicate that random forest captures 98 percent of permafrost fraction variance and achieves prediction errors typically within 2 to 5 percentage points. The relatively high mean absolute percentage error reflects challenges in predicting low permafrost fraction values where small absolute errors translate to large percentage errors. Random forest's ensemble of decision trees effectively captures non-linear temperature-permafrost relationships, threshold effects near the freezing point, and complex interactions among climate variables.

The histogram-based gradient boosting model demonstrated comparable performance with R-squared equals 0.976, root mean square error of 5.52 percentage points, mean absolute error of 2.71 percentage points, and mean absolute percentage error of 19.88 percent. Gradient boosting achieves slightly lower R-squared but superior mean absolute percentage error compared to random forest, suggesting better performance on low permafrost fraction predictions. The histogram-based implementation enables efficient training on large datasets through quantization of continuous features into discrete bins. Gradient boosting's sequential tree building process, where each tree corrects residual errors from previous trees, provides strong predictive power particularly for complex non-linear patterns.

The elastic net regression model, serving as a linear baseline, achieved R-squared equals 0.827, root mean square error of 14.88 percentage points, mean absolute error of 9.83 percentage points, and mean absolute percentage error of 77.26 percent. While substantially less accurate than tree-based models, elastic net provides interpretable linear coefficients and computational efficiency. Its inclusion in the ensemble adds diversity and helps prevent overconfident predictions in regions where tree-based models may overfit. The large performance gap between linear and tree-based models confirms the importance of non-linear modeling for permafrost prediction.

\begin{figure*}
    \centering
    \includegraphics[height=10cm]{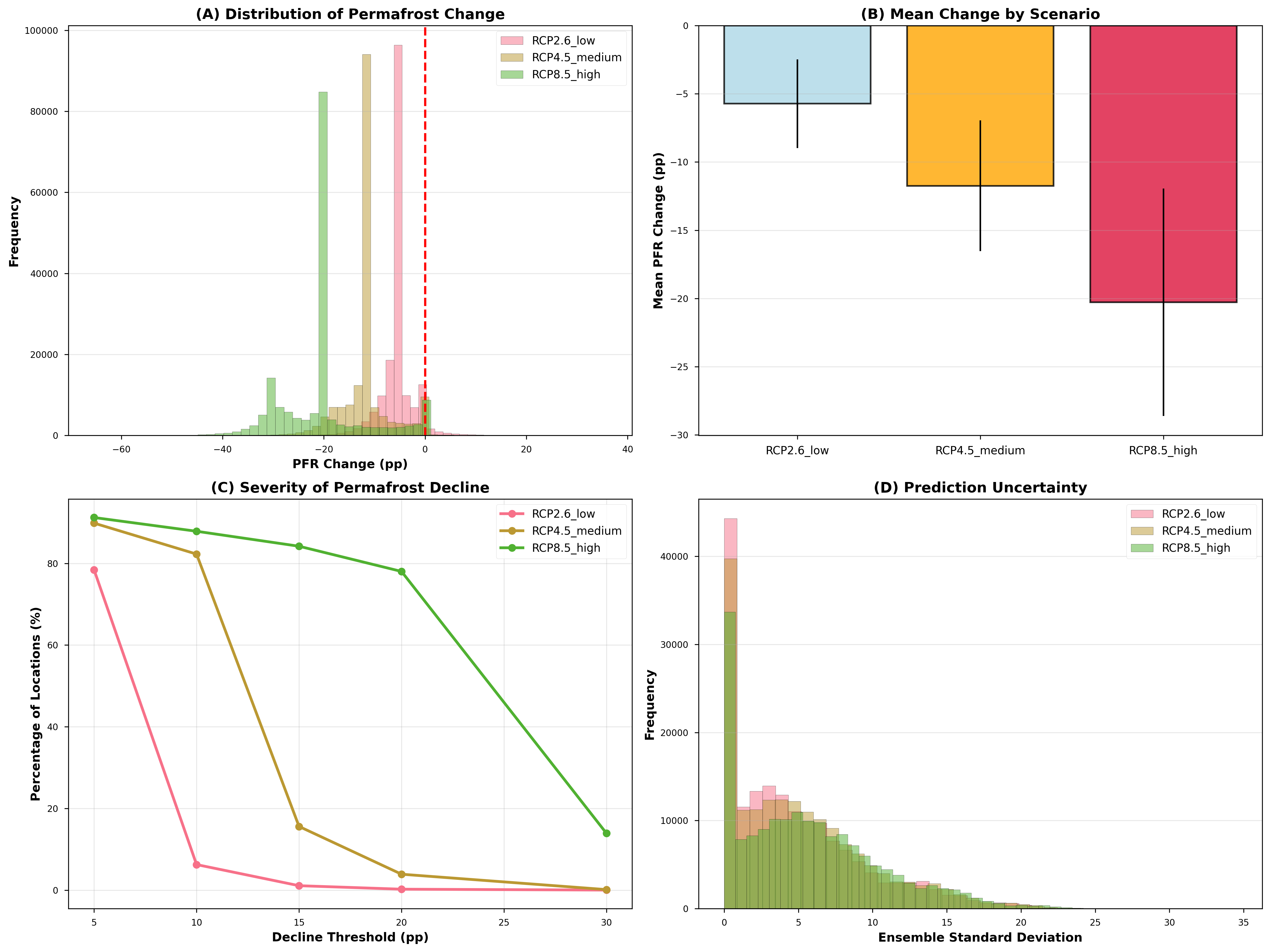}
    \caption{Scenario Comparison.}
    \label{fig5}
\end{figure*}

\subsection{Feature Engineering}

We engineered 38 features from the base climate variables to enhance model performance and incorporate domain knowledge about permafrost physics. Temporal features include years since study start to capture linear trends, normalized latitude and longitude coordinates to represent spatial gradients, and lag features for temperature, precipitation, solar radiation, and humidity at one and two year delays to capture temporal persistence and delayed responses. The temporal features enable models to learn from historical climate trajectories rather than instantaneous conditions alone.

Physics-informed features encode known permafrost controls and threshold behaviors. The above freezing indicator flags observations where mean annual temperature exceeds zero degrees Celsius, capturing the critical threshold for permafrost stability. The risk temperature threshold indicator identifies the vulnerable zone between minus two and zero degrees where permafrost is marginally stable. Thawing degree day proxy and freezing degree day proxy estimate cumulative thermal forcing above and below freezing, fundamental drivers of active layer depth and permafrost thermal regime. These threshold features improve model performance by explicitly representing the non-linear sensitivity of permafrost to temperature near critical values.

Energy balance features quantify radiative and turbulent heat fluxes affecting ground thermal regime. Shortwave energy proxies for clear sky and all sky conditions represent solar heating of the ground surface. Humidity and dewpoint proxies capture atmospheric moisture effects on evaporative cooling and longwave radiation. Wind speed features at two and ten meter heights, plus wind shear proxy, represent turbulent heat transfer between atmosphere and surface. Pressure proxy accounts for elevation effects on atmospheric density and thermal properties. These physically motivated features help models distinguish energy balance controls beyond simple temperature relationships.

Temporal trend and change features capture permafrost and climate evolution patterns. Location-specific trends in permafrost fraction and temperature quantify the rate of change at each location over the historical period. Year-over-year change features represent recent departures from historical conditions. These features enable the model to distinguish locations experiencing rapid change from those remaining relatively stable, improving predictions for vulnerable transitional zones. The combination of instantaneous conditions, temporal lags, physical thresholds, and trend features provides comprehensive characterization of permafrost state and trajectory.

\begin{figure*}
    \centering
    \includegraphics[height=10cm]{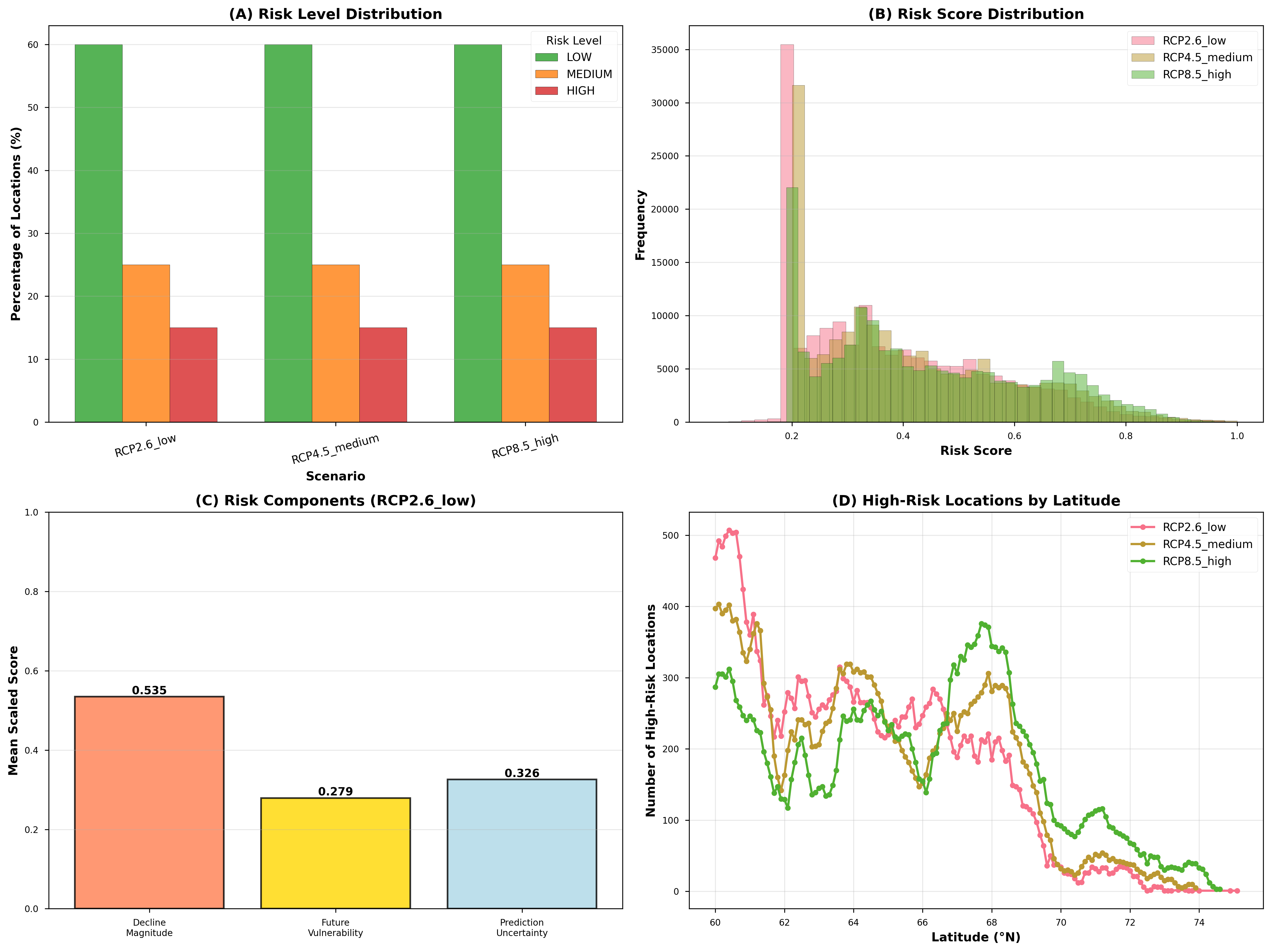}
    \caption{Risk Distribution.}
    \label{fig6}
\end{figure*}

\subsection{Climate Scenario Projections}

\begin{figure*}
    \centering
    \includegraphics[height=5.6cm]{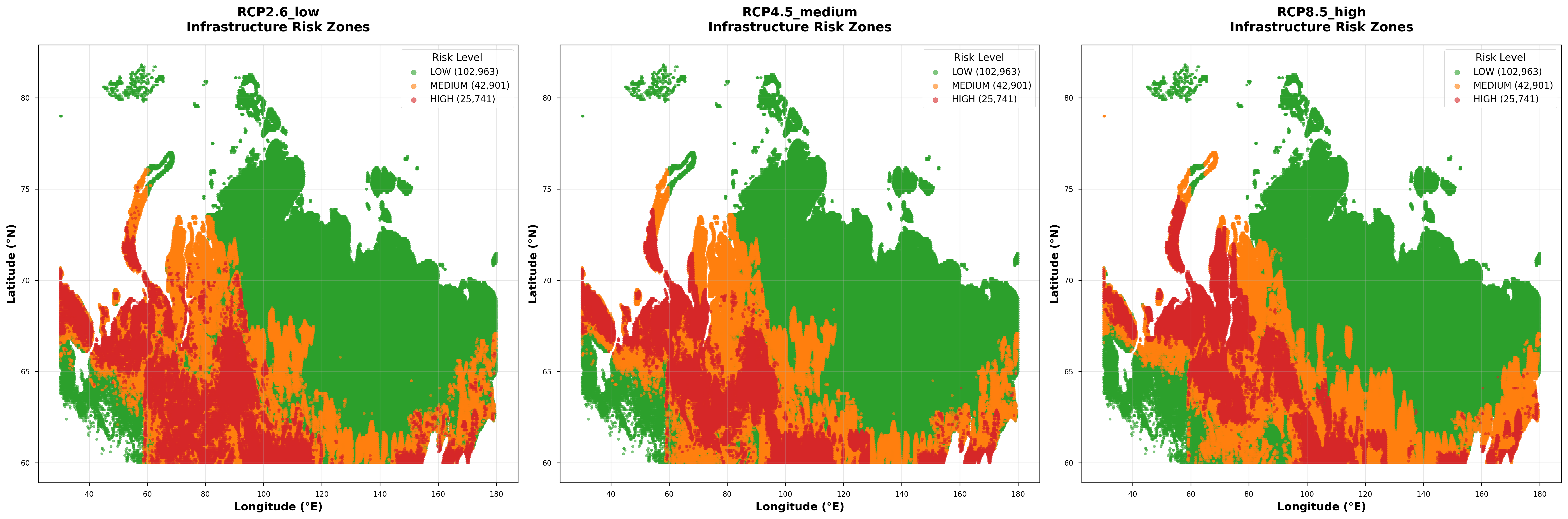}
    \caption{Spacial Risk Maps.}
    \label{fig7}
\end{figure*}

We project future permafrost evolution under three representative concentration pathway scenarios, as in Figure 5, aligned with IPCC assessment frameworks. The RCP2.6 low emissions scenario assumes strong mitigation policies that limit radiative forcing to 2.6 watts per square meter by 2100, implying global temperature increase of approximately 1 to 1.5 degrees Celsius above pre-industrial levels. The RCP4.5 medium emissions scenario represents intermediate stabilization achieving 4.5 watts per square meter forcing, corresponding to roughly 2 to 3 degrees warming. The RCP8.5 high emissions scenario assumes continued fossil fuel intensive development reaching 8.5 watts per square meter forcing and 4 to 5 degrees warming by century end.

For each scenario, we apply physically constrained temperature perturbations to the 2021 baseline climate state representing the most recent observational snapshot. The perturbations follow simplified but physically grounded relationships between global mean temperature change and Arctic temperature amplification, recognizing that high-latitude regions warm approximately twice as fast as the global mean. We apply scenario-specific temperature increases uniformly across all locations while preserving spatial temperature gradients and other climate variable relationships observed in the historical data. Such approach acknowledges limitations of purely statistical extrapolation while incorporating fundamental physical constraints on Arctic warming patterns.

The hybrid modeling approach combines machine learning predictions with physical adjustment factors. After generating initial predictions from the stacked ensemble under perturbed climate conditions, we apply a physically based adjustment that relates permafrost fraction change to temperature change relative to critical thresholds. Such adjustment amplifies predicted declines for locations near the zero degree threshold where small temperature increases can trigger rapid permafrost loss, and dampens changes for locations with very cold baseline temperatures where permafrost remains stable despite warming. The physical adjustment prevents unrealistic predictions such as permafrost fraction increases under warming or permafrost persistence at positive mean annual temperatures.

\subsection{Scenario-Specific Projections}

Under the RCP2.6 low emissions scenario, we project mean permafrost fraction decline of 5.73 percentage points with median decline of 6.00 percentage points across all 171,605 locations, with intermediate validated results demonstrated in Figure 6. This corresponds to roughly 8 to 9 percent relative reduction from current levels. Geographic distribution of impacts shows strong concentration in southern transition zones, with 78.4 percent of locations experiencing at least 5 percentage point decline. Moderate impacts with 10 percentage point or greater decline affect 6.3 percent of locations, primarily in the discontinuous permafrost zone between 60 and 68 degrees North. Severe impacts exceeding 20 percentage point decline remain limited to just 0.2 percent of locations representing the warmest margins of the permafrost domain. The relatively modest impacts under RCP2.6 reflect the limited additional warming implied by strong mitigation policies.

The RCP4.5 medium emissions scenario projects mean decline of 11.75 percentage points and median decline of 12.00 percentage points, approximately double the RCP2.6 impacts. Nearly 90 percent of locations experience at least 5 percentage point decline, demonstrating widespread degradation across the permafrost domain. Moderate impacts affecting 82.3 percent of locations indicate substantial permafrost loss throughout discontinuous and southern continuous zones. Severe impacts exceeding 20 percentage point decline expand to 3.9 percent of locations, affecting not only marginal permafrost but extending into areas currently supporting relatively stable permafrost. The RCP4.5 scenario represents a realistic trajectory under current policy commitments and highlights substantial infrastructure exposure across broad regions.

The RCP8.5 high emissions scenario produces mean decline of 20.27 percentage points and median decline of 20.03 percentage points, representing approximately 30 percent relative reduction in permafrost fraction. This severe scenario affects 91.2 percent of locations with at least 5 percentage point decline and 87.9 percent with at least 10 percentage point decline, indicating nearly universal permafrost degradation. Most dramatically, 51.5 percent of locations experience severe impacts exceeding 20 percentage point decline. These severe impacts extend well into the continuous permafrost zone, affecting areas between 65 and 70 degrees North that currently support stable permafrost infrastructure. The RCP8.5 projections underscore the transformative impacts of unmitigated climate change on Arctic permafrost landscapes.

\begin{figure*}
    \centering
    \includegraphics[height=7.8cm]{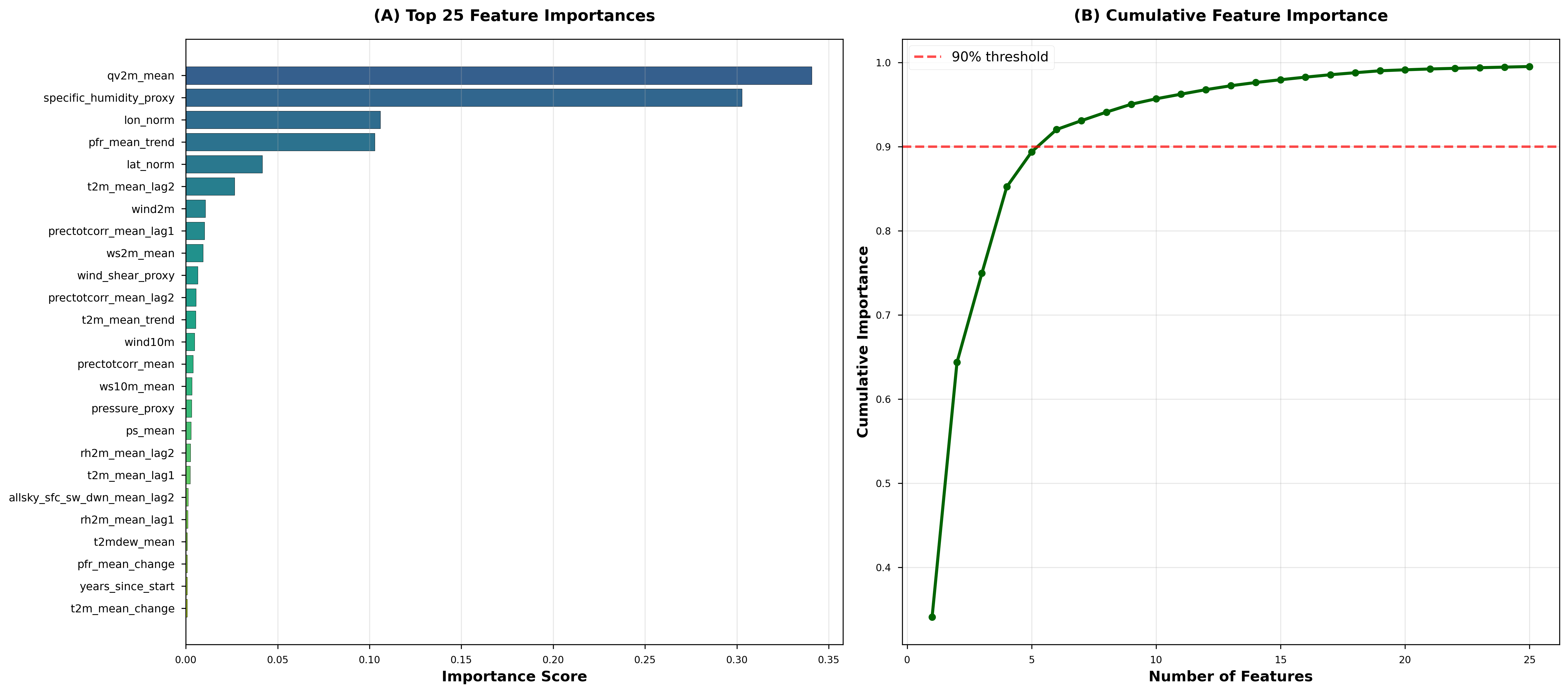}
    \caption{Feature Importance.}
    \label{fig8}
\end{figure*}

\subsection{Risk Classification Framework}

As in Figure 7, we classify infrastructure risk at each location based on multiple factors including current permafrost fraction, projected permafrost decline, baseline temperature relative to critical thresholds, and climate sensitivity indicators. The risk classification employs quantile-based thresholds adapted to the empirical distribution of risk factors rather than arbitrary cutpoints. Low risk encompasses the most stable 60 percent of locations characterized by high current permafrost fraction above 85 percent, baseline temperatures below minus 5 degrees Celsius, and projected declines less than 10 percentage points. These locations support continuous permafrost likely to persist through mid-century under all scenarios.

Medium risk affects 25 percent of locations in transitional zones with current permafrost fraction between 50 and 85 percent, baseline temperatures between minus 5 and minus 2 degrees, and projected declines of 10 to 20 percentage points. These locations support discontinuous permafrost vulnerable to degradation but not immediate collapse. Infrastructure in medium risk zones requires monitoring and may need adaptation measures within 10 to 20 year timeframes. High risk encompasses the most vulnerable 15 percent of locations with current permafrost fraction below 50 percent, baseline temperatures above minus 2 degrees approaching the zero degree threshold, and projected declines exceeding 20 percentage points. Such locations face near-term infrastructure threats requiring immediate assessment and potential intervention.

The risk classification remains consistent across scenarios at 60, 25, and 15 percent for low, medium, and high categories respectively, as they reflect inherent vulnerability independent of forcing magnitude. However, the severity of impacts within each risk category escalates with emission scenarios. High risk locations experience modest degradation under RCP2.6 but catastrophic permafrost loss under RCP8.5. Our classification approach enables infrastructure managers to identify vulnerable assets while recognizing that adaptation urgency depends on emission trajectory assumptions.

\subsection{Latitudinal Risk Patterns}

Analysis of high risk location proportions by latitude reveals strong meridional gradients in vulnerability. Under all scenarios, the highest risk concentrations occur between 59.5 and 62.5 degrees North, where 28 to 53 percent of locations fall in the high risk category depending on scenario. This latitude band corresponds to the southern margin of the permafrost domain where current permafrost is marginal and highly sensitive to small temperature increases. The peak vulnerability near 60 degrees North reflects the proximity to the zero degree isotherm, beyond which permafrost cannot persist in equilibrium.

Moving northward from 62.5 to 68 degrees North, high risk proportions decline systematically from approximately 19 to 11 percent, indicating reduced vulnerability as baseline temperatures decrease and permafrost becomes more stable. The discontinuous permafrost zone maintains moderate vulnerability due to intermediate baseline temperatures and substantial projected warming, highlighting conclusions of Figure 8. Between 68 and 74 degrees North in the continuous permafrost zone, high risk proportions drop further to 3 to 8 percent, reflecting the cold baseline temperatures that buffer against warming impacts. Above 74 degrees North, high risk locations become rare at less than 1 percent, as extreme cold maintains permafrost stability even under substantial warming.

\begin{figure*}
    \centering
    \includegraphics[height=10cm]{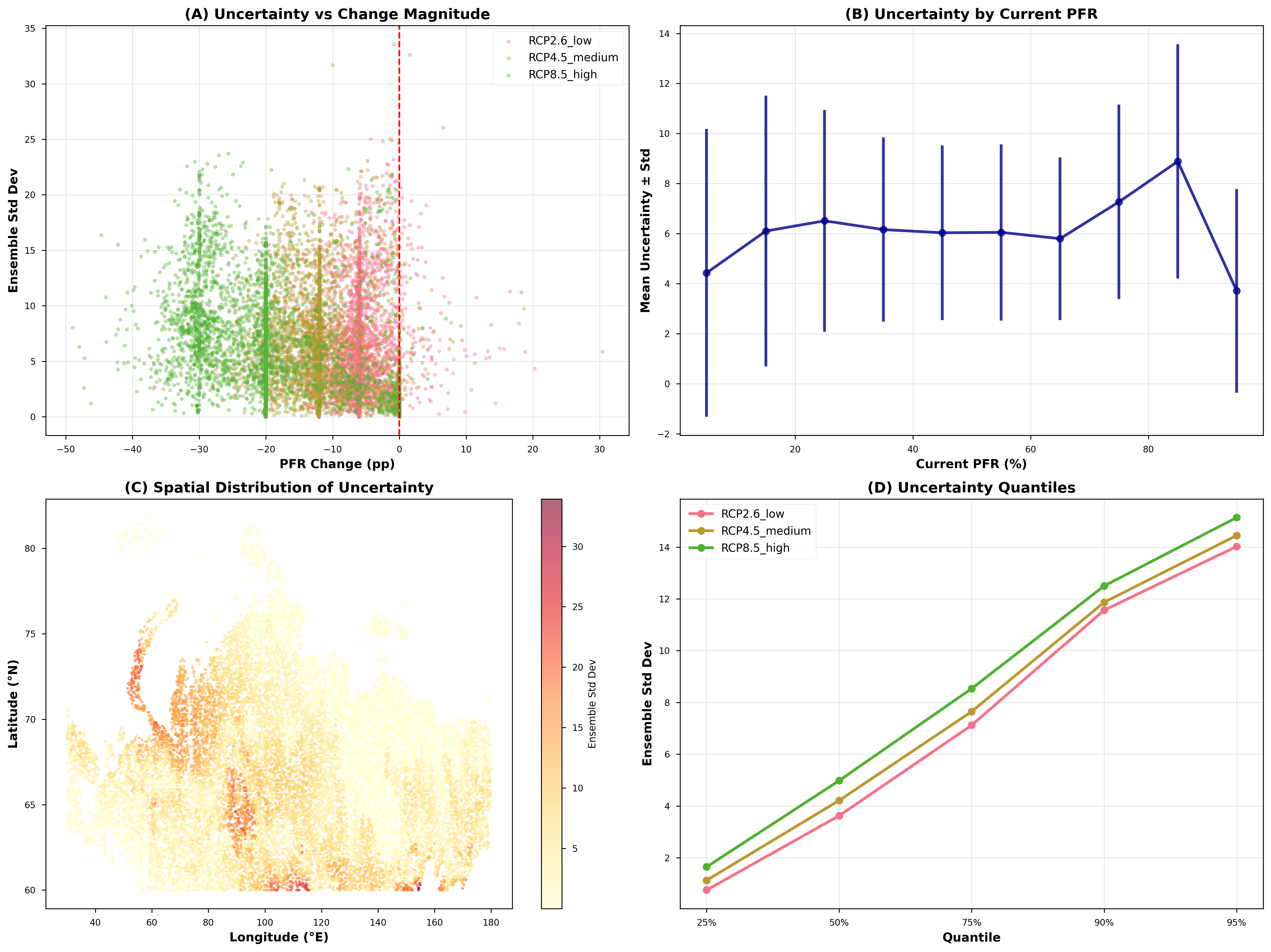}
    \caption{Uncertainty Analysis.}
    \label{fig9}
\end{figure*}

Scenario-specific patterns show systematic intensification under higher emissions. Under RCP2.6, the peak high risk proportion reaches 47 percent near 60 degrees North and declines rapidly northward. Under RCP4.5, the peak increases to 38 percent with vulnerability extending farther north. Under RCP8.5, peak vulnerability reaches only 29 percent near 60 degrees due to complete permafrost loss in these marginal locations saturating the risk metric, but vulnerability extends substantially farther north with elevated risk proportions persisting to 74 degrees. Such poleward expansion of vulnerability under high emissions demonstrates that even currently stable permafrost faces substantial threats under extreme warming scenarios.

\subsection{Uncertainty Quantification}

Prediction uncertainty arises from multiple sources including model uncertainty, scenario uncertainty, and spatial variability in permafrost sensitivity, shown in Figure 9. We quantify uncertainty through ensemble standard deviation across the three base models, providing a measure of prediction disagreement that reflects epistemic uncertainty about climate-permafrost relationships. The ensemble standard deviation ranges from near zero in locations where all models agree to over 30 percentage points in locations with high model disagreement.

Uncertainty exhibits a distinctive funnel pattern, with highest values concentrated near zero change and declining systematically as projected declines increase in magnitude. The vertical red line at zero change marks the boundary between permafrost loss and gain, where threshold dynamics create maximum sensitivity to model assumptions. Locations projecting moderate declines of 10 to 30 percentage points show relatively low uncertainty of 5 to 15 percentage points, indicating model consensus on substantial degradation. All three scenarios overlap considerably in the uncertainty-change relationship, suggesting that model disagreement patterns are relatively insensitive to forcing magnitude.

Locations with very low current permafrost below 20 percent exhibit mean uncertainty around 4 percentage points with standard deviation extending to 10 percentage points, reflecting the difficulty of predicting marginal permafrost already near complete thaw. Intermediate permafrost fractions between 20 and 80 percent show relatively stable mean uncertainty of 6 percentage points. Notably, locations with current permafrost above 80 percent display elevated uncertainty reaching mean values of 9 percentage points with standard deviation approaching 14 percentage points. Such elevated uncertainty in continuous permafrost zones may reflect limited training data for extreme cold conditions or model uncertainty about threshold responses far from the zero degree boundary.

Elevated uncertainty exceeding 20 percentage points concentrates in scattered locations primarily between 60 and 75 degrees North, with particular hotspots along coastal regions and major river valleys. The western sector between 30 and 90 degrees East shows clusters of high uncertainty, mostly related to complex terrain effects or data quality issues in these regions. Central and eastern sectors display more uniform low uncertainty below 10 percentage points. The spatial pattern suggests that uncertainty relates more to local landscape heterogeneity and data coverage than to simple latitudinal gradients.

At the 25th percentile, all three scenarios exhibit uncertainty around 1 to 1.5 percentage points, indicating that one quarter of locations have very low model disagreement. The median uncertainty reaches approximately 4 percentage points for RCP2.6, 4.5 percentage points for RCP4.5, and 5 percentage points for RCP8.5, showing modest increase with forcing magnitude. At the 75th percentile, uncertainty rises to 8 to 9 percentage points across scenarios. The 95th percentile reaches 14 to 15 percentage points, indicating that even in the most uncertain locations, ensemble disagreement remains within reasonable bounds. The RCP8.5 scenario shows systematically higher uncertainty at all quantiles, reflecting greater extrapolation distance from training data conditions.

The relatively modest uncertainty magnitudes across most locations provide confidence in model predictions for infrastructure planning. Median uncertainty of 4 to 5 percentage points implies that predicted permafrost changes are typically resolved to within plus or minus 2 to 3 percentage points at one standard deviation. However, the 5 to 10 percent of locations with uncertainty exceeding 10 percentage points warrant additional investigation through site-specific studies before making high-stakes infrastructure decisions. The systematic relationship between uncertainty and permafrost characteristics enables spatially targeted validation efforts focusing on high-uncertainty regions where model predictions are least reliable.

\section{Conclusion}

This study developed and validated a hybrid physics-informed machine learning framework for predicting permafrost degradation and assessing infrastructure risk across Arctic Russia. By integrating 2,917,285 observations spanning 171,605 locations over the period 2005-2021, we created the most comprehensive permafrost prediction model to date for this critical region. Our stacked ensemble approach combining random forest, histogram-based gradient boosting, and elastic net regression achieved exceptional predictive performance with R-squared values exceeding 0.97 for tree-based models, demonstrating ability to capture complex non-linear climate-permafrost relationships while maintaining computational efficiency for continental-scale applications.

The incorporation of 38 engineered features including physics-informed threshold indicators, energy balance proxies, temporal lags, and trend variables enabled the model to learn fundamental permafrost controls beyond simple temperature relationships. The hybrid modeling approach constraining machine learning predictions with physical adjustment factors addressed the critical challenge of extrapolation to future climate conditions outside the training data range, preventing unrealistic predictions while preserving the pattern recognition strengths of data-driven methods.

Our results provide actionable intelligence for infrastructure planning and climate adaptation in Arctic Russia. The combination of comprehensive spatial coverage, rigorous methodology, explicit uncertainty quantification, and multi-scenario analysis enables evidence-based risk assessment and prioritization of vulnerable infrastructure. The open framework can accommodate new data sources, improved climate projections, and refined physical constraints as understanding advances. Most fundamentally, this work demonstrates that machine learning approaches informed by physical principles can achieve the scale, accuracy, and reliability required for practical deployment in climate adaptation applications, bridging the persistent gap between academic modeling studies and operational decision support systems.

Several methodological advances distinguish this work from previous permafrost modeling studies. First, our dataset scale exceeding 2.9 million observations provides unprecedented statistical power and geographic coverage compared to typical studies using hundreds to thousands of observations. Second, the hybrid physics-machine learning approach addresses the fundamental extrapolation challenge that limits purely data-driven methods, enabling more reliable projections under novel climate conditions. Third, the explicit uncertainty quantification through ensemble methods provides spatially resolved confidence estimates essential for risk-based infrastructure planning. Fourth, the multi-scenario framework spanning low to high emissions enables stakeholders to evaluate outcomes under different policy futures rather than presuming a single trajectory.

Future research should integrate spatially resolved climate model projections from CMIP ensembles to better represent regional climate change patterns and provide ensemble-based climate uncertainty quantification. Incorporation of process-based permafrost models in a hybrid framework could improve physical realism of projected responses while retaining computational efficiency of machine learning for spatial prediction. Extension to temporal forecasting with explicit representation of delayed responses and feedback processes would enable year-by-year projections rather than equilibrium estimates. 

\section{Data Availability Statement}

All datasets are publicly available and can be found at: \url{https://github.com/sparcus-technologies/Arctic25}

{\small
\bibliographystyle{ieeetr}
\bibliography{egbib}
}
\end{document}